\documentclass[10pt,twocolumn,letterpaper]{article}

\usepackage[pagenumbers]{cvpr}

\usepackage{multirow}
\usepackage{colortbl}
\usepackage{tabularx}
\definecolor{MemAnchorWorldRow}{RGB}{252,235,218}

\newcommand{\MemAnchorQualitative}{%
\includegraphics[width=0.91\textwidth]{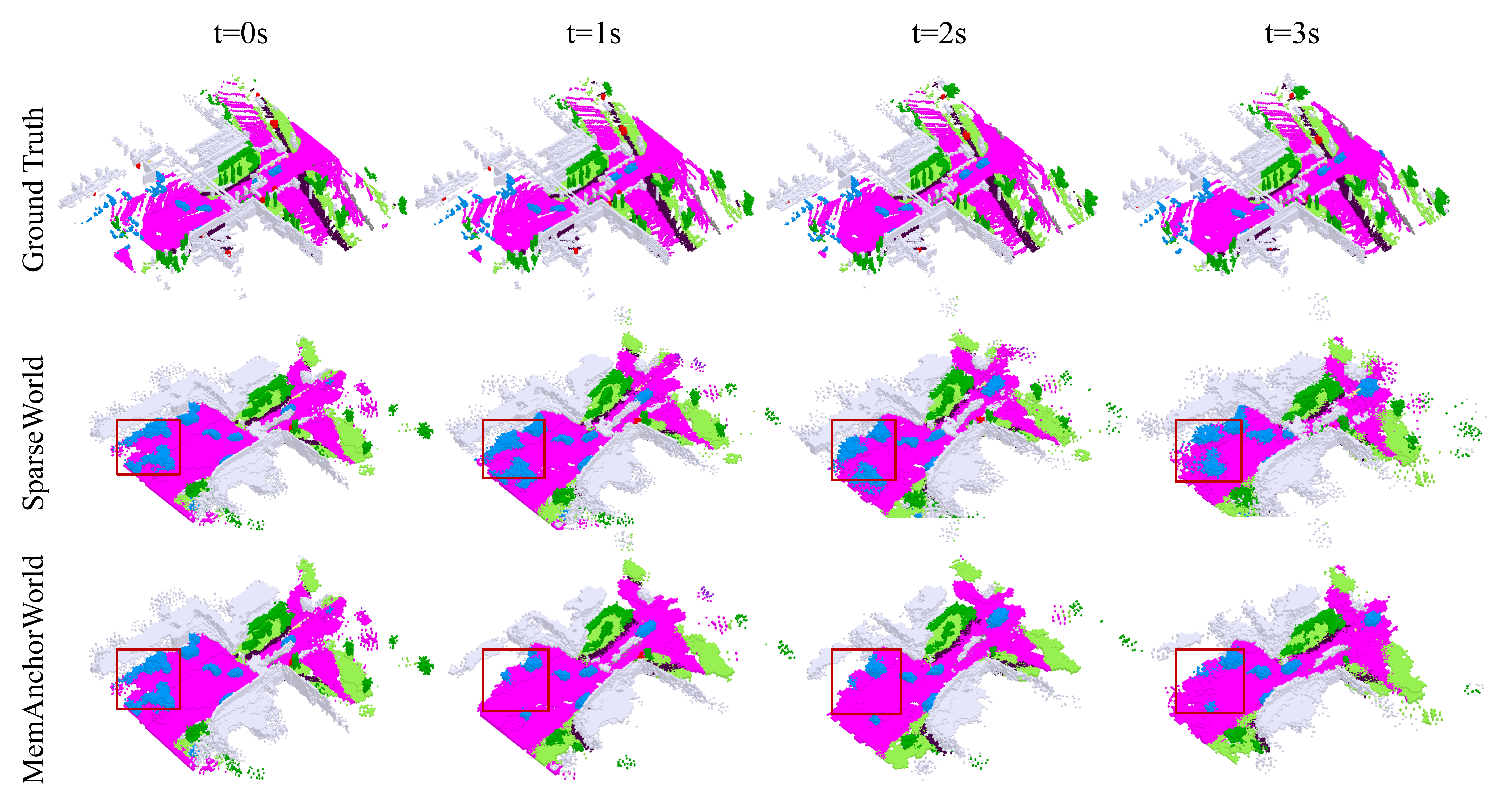}%
}
\newcommand{\MemAnchorOcclusion}{%
\includegraphics[width=0.85\textwidth]{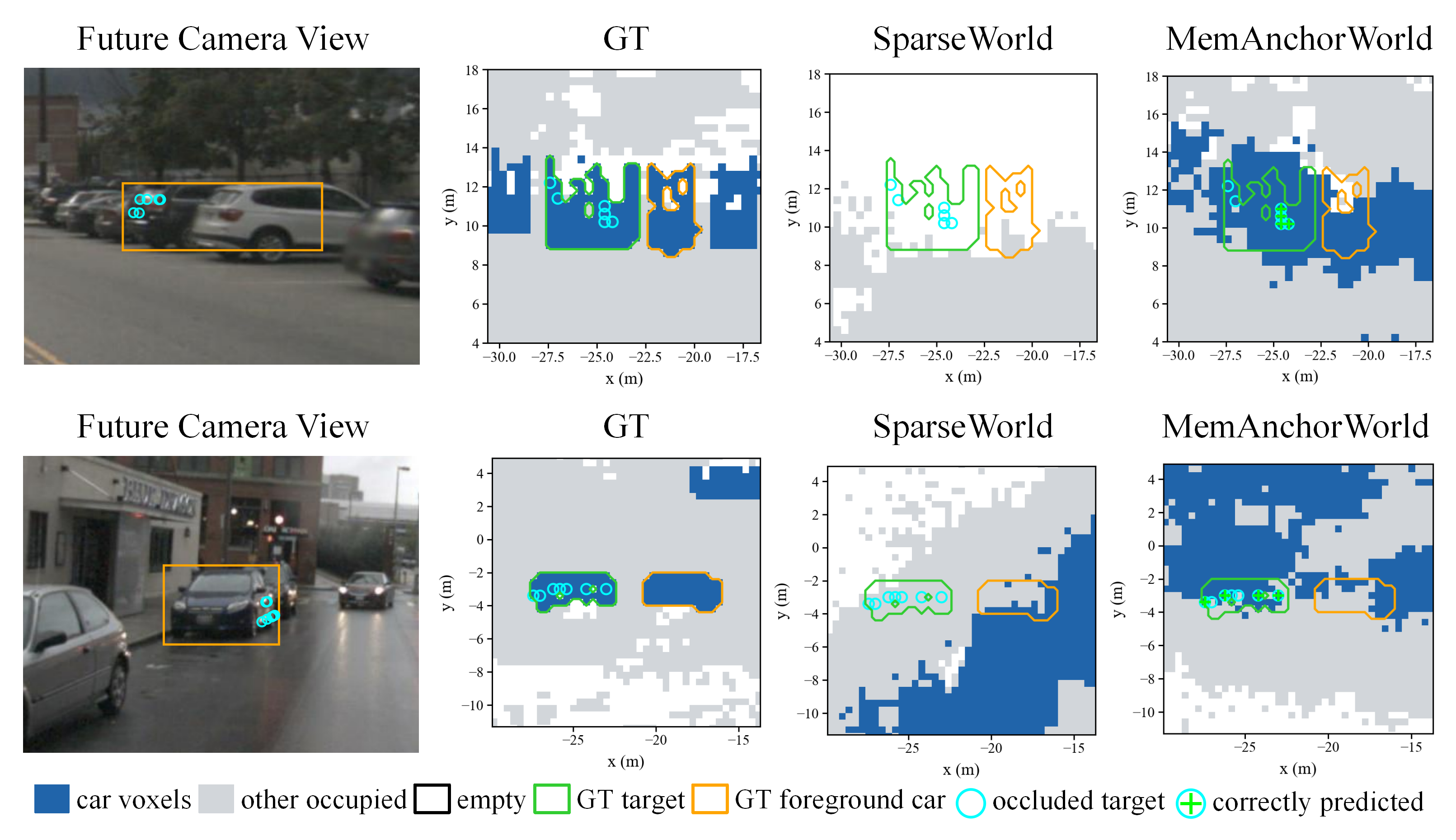}%
}

\definecolor{cvprblue}{rgb}{0.21,0.49,0.74}
\usepackage[pagebackref,breaklinks,colorlinks,allcolors=cvprblue]{hyperref}

\def\confName{CVPR}
\def\confYear{2027}

\title{MemAnchorWorld: A Sparse World Model with Horizon-Conditioned Memory Retrieval for Autonomous Driving}

\author{%
Xinyang Jia$^{1}$ \quad
Yuqi Zhang$^{2}$\thanks{Corresponding authors.} \quad
Jiakai Wang$^{2}$\footnotemark[1] \quad
Yahu Yang$^{2}$\\
Guandong Pan$^{3}$ \quad
Chao Tong$^{1}$ \quad
Xianglong Liu$^{1,2}$\\[0.4em]
{\normalsize $^{1}$School of Computer Science and Engineering, Beihang University}\\
{\normalsize $^{2}$Zhongguancun Laboratory}\\
{\normalsize $^{3}$School of Artificial Intelligence, Beihang University}
}

\hypersetup{
  pdftitle={MemAnchorWorld: A Sparse World Model with Horizon-Conditioned Memory Retrieval for Autonomous Driving},
  pdfauthor={Xinyang Jia, Yuqi Zhang, Jiakai Wang, Yahu Yang, Guandong Pan, Chao Tong, Xianglong Liu}
}

\begin{document}
\maketitle
\begin{abstract}

Sparse world models for 4D semantic occupancy forecasting represent scenes with a small set of evolving queries to efficiently predict future geometry and semantics for autonomous driving. However, existing sparse world models face a \textbf{bottleneck in using historical temporal evidence for long-horizon forecasting}. They rely on scene states to carry historical information implicitly, allowing errors to propagate and accumulate across successive forecasting steps and causing geometric and semantic deviations. To address this limitation, we present \textbf{MemAnchorWorld}, a sparse occupancy world model that explicitly retrieves relevant historical geometric and semantic evidence for 4D semantic occupancy forecasting. Inspired by the von Restorff effect, we introduce a \textbf{forecast-time-conditioned hierarchical dual-path memory-anchor retrieval mechanism}. Guided by the effective age and semantic compatibility of historical information, it combines query-level candidate selection with point-level memory-anchor retrieval to provide historical evidence for sparse points at different forecast times. To better use memory anchors and existing features, \textbf{a point-level residual adapter jointly gated by the task and forecast time} separately controls residual corrections to semantics, occupancy confidence, and spatial positions.  On Occ3D-nuScenes, MemAnchorWorld achieves an average mIoU of 13.55\% and an average IoU of 22.84\% across the 1–3 s forecasting horizons, attaining state-of-the-art performance on both metrics. Project page: \url{https://memanchorworld.github.io/}.


\end{abstract}
    
\section{Introduction}
\label{sec:intro}

Vision-centric autonomous driving has been widely studied, with substantial progress in scene perception~\cite{Huang_2023_CVPR,Cao_2022_MonoScene,Li_2023_VoxFormer,Wei_2023_ICCV}, motion prediction~\cite{Hu_2021_FIERY,Zhang_2022_BEVerse,Sun_2025_SparseDrive,Fang_2023_TBPFormer}, and trajectory planning~\cite{Hu_2022_STP3,Hu_2023_CVPR,Jiang_2023_ICCV,Sun_2025_SparseDrive}. More recent methods encode scene information in learnable sparse queries~\cite{Liu_2024_SparseOcc,Wang_2024_OPUS,Lin_2023_Sparse4Dv2,Sun_2025_SparseDrive}, using adaptive position updates and feature interactions to capture local geometry and semantics. Related compact representations support streaming scene aggregation~\cite{Zuo_2025_GaussianWorld,Park_2026_S2GO,Lin_2023_Sparse4Dv2,Boeder_2025_GaussianFlowOcc} and future occupancy forecasting~\cite{Xu_2025_EfficientOCF,Dang_2026_SparseWorld,Zhu_2026_DLWM,Feng_2025_GWM}.


\begin{figure}[t]
\centering
\includegraphics[width=\columnwidth]{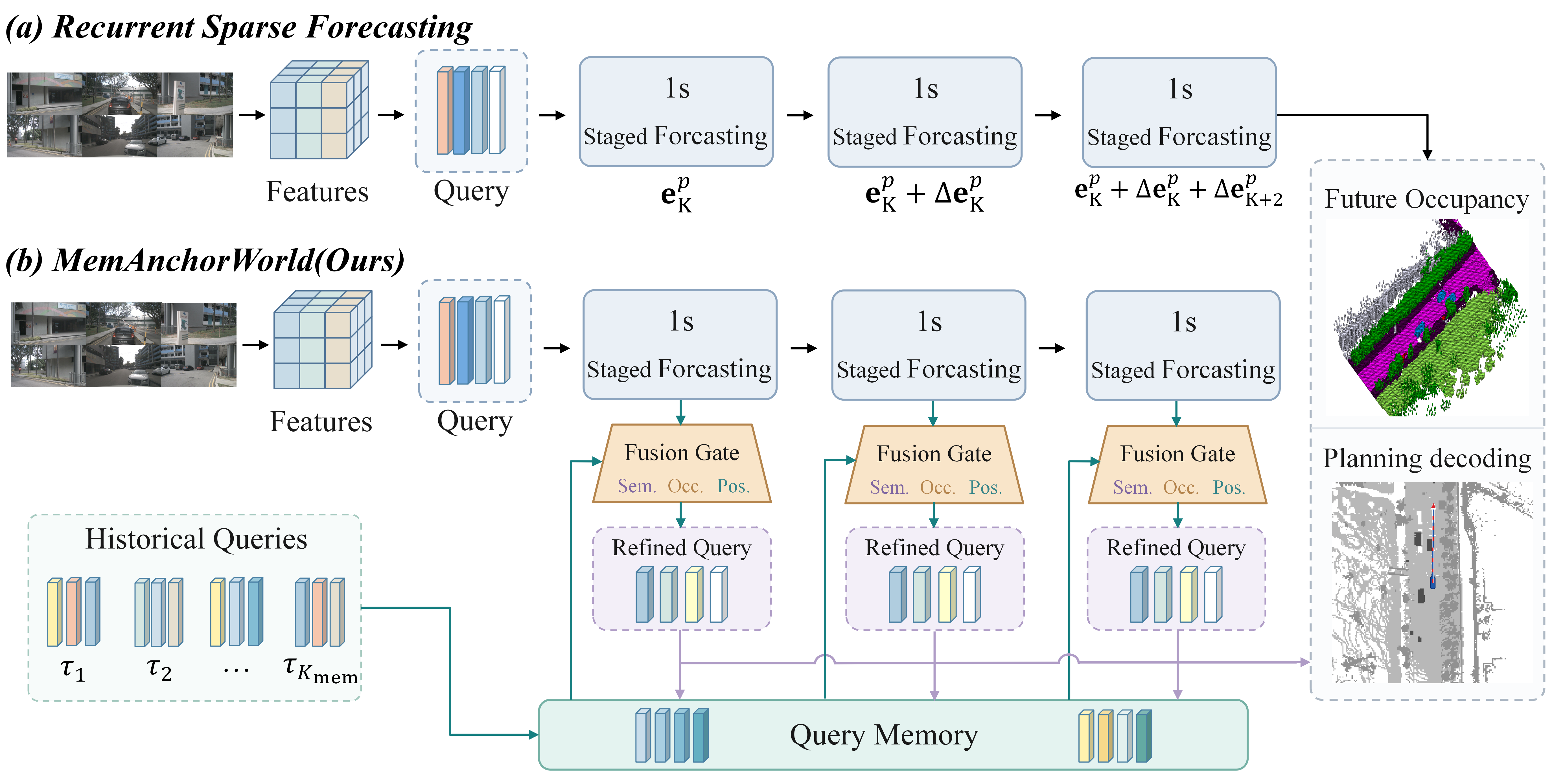}
\caption{Comparison of recurrent sparse occupancy forecasting and MemAnchorWorld. Recurrent forecasting updates scene queries through successive steps. MemAnchorWorld retrieves memory anchors from a separate historical memory and uses task-specific gates to refine the outputs at each forecast horizon.}
\label{fig:method_comparison}
\end{figure}

However, existing sparse world models still face a temporal evidence bottleneck in long-horizon forecasting. As shown in \cref{fig:method_comparison}(a), occupancy world models that predict future scenes through autoregressive or state-conditioned updates~\cite{Zheng_2024_OccWorld,Li_2025_PreWorld,Dang_2026_SparseWorld,Yang_2026_GenieDrive} primarily rely on the recurrent state to carry historical information. When local geometric and semantic details are not sufficiently preserved during encoding, subsequent predictions struggle to recover them, and errors may propagate and accumulate during rollout.

To address this issue, as shown in \cref{fig:method_comparison}(b), we propose MemAnchorWorld, a sparse world model for 4D semantic occupancy forecasting. Inspired by the von Restorff effect~\cite{vonRestorff_1933_IsolationEffect}, we introduce a forecast-time-conditioned hierarchical dual-path memory-anchor retrieval mechanism. Considering the effective age and semantic uncertainty of historical information, this mechanism first selects candidate historical queries through query-level geometric--semantic dual-path coarse retrieval. It then performs point-level retrieval among their associated points to select relevant historical points as memory anchors for each predicted point, providing local geometric and semantic evidence for future forecasting. To adaptively fuse evidence from memory anchors with the current prediction representation, we introduce a point-level residual adapter jointly gated by task and forecast time, which separately modulates residual corrections to semantics, occupancy confidence, and spatial positions. Furthermore, to improve consistency between point-level supervision and voxel-based evaluation, we introduce voxel-center-aligned sparse differentiable joint supervision.


We evaluate MemAnchorWorld on camera-based 4D occupancy forecasting and downstream planning using Occ3D-nuScenes~\cite{Tian_2023_Occ3D,Caesar_2020_nuScenes}. In our evaluation, MemAnchorWorld achieves an average mIoU of 13.55\% and an average IoU of 22.84\% across the 1--3 s forecasting horizons. By retrieving memory anchors at each forecast horizon, MemAnchorWorld mitigates the impact of errors accumulated through recurrent state propagation on future occupancy predictions. Our contributions are as follows:


\begin{itemize}

    \item We propose a forecast-time-conditioned hierarchical dual-path memory-anchor retrieval mechanism combining age- and semantics-aware query selection with point-level memory-anchor retrieval.

    \item We develop a point-level residual adapter jointly gated by task and forecast time and introduce voxel-center-aligned differentiable joint supervision to more closely align point-level supervision with voxel-based evaluation.

    \item On Occ3D-nuScenes, MemAnchorWorld achieves an average mIoU of 13.55\% and an average IoU of 22.84\% in the current evaluation setting, outperforming SparseWorld by 0.35 and 0.81 percentage points, respectively.
\end{itemize}


\section{Related Work}
\label{sec:related_work}

\subsection{Camera-Based Semantic Occupancy}

Camera-based methods infer scene geometry and semantics from visual cues, offering lower sensor costs and flexible deployment~\cite{Li_2023_VoxFormer,Wei_2023_ICCV,Li_2025_PreWorld,Tong_2023_OccNet,Ma_2024_COTR}. MonoScene~\cite{Cao_2022_MonoScene} recovers scene geometry and semantics from a monocular image through 2D-to-3D feature projection. VoxFormer~\cite{Li_2023_VoxFormer} uses depth estimation to construct sparse voxel queries for visible occupied regions, followed by feature propagation for scene completion. For surround-view scenes, SurroundOcc~\cite{Wei_2023_ICCV} aggregates multi-camera features through multi-scale 2D--3D attention to construct a 3D volumetric representation, whereas TPVFormer~\cite{Huang_2023_CVPR} encodes 3D space using three mutually orthogonal planes to reduce computation and storage costs. Other camera-based approaches include voxel occupancy in OccNet~\cite{Tong_2023_OccNet}, compact 3D encoding in COTR~\cite{Ma_2024_COTR}, video-based self-supervision in SelfOcc~\cite{Huang_2024_SelfOcc}, and panoptic occupancy in PanoOcc~\cite{Wang_2024_PanoOcc}.


To reduce redundant computation in unoccupied regions, later studies explore sparse scene representations. SparseOcc~\cite{Liu_2024_SparseOcc} combines sparse 3D reconstruction with mask-guided sampling. OPUS~\cite{Wang_2024_OPUS} directly predicts occupied point locations and semantic classes from learnable queries. Other methods use parameterized geometric primitives to describe local space. GaussianFormer-2~\cite{Huang_2025_GaussianFormer2} combines probabilistic Gaussian superposition with semantic mixture modeling to make scene representations more compact. For temporal modeling, GaussianWorld~\cite{Zuo_2025_GaussianWorld} decomposes scene evolution into ego-motion alignment, dynamic-object motion, and completion of newly observed regions, and updates its Gaussian scene state accordingly.


\subsection{4D Occupancy Forecasting and World Models}

4D occupancy forecasting predicts future scene occupancy from historical observations~\cite{Khurana_2023_PointForecast}. Cam4DOcc~\cite{Ma_2024_Cam4DOcc} establishes a benchmark for camera-based 4D occupancy forecasting, while UniOcc~\cite{Wang_2025_UniOcc} provides a unified benchmark for occupancy forecasting and current-frame prediction. EfficientOCF~\cite{Xu_2025_EfficientOCF} decomposes 3D occupancy into bird's-eye-view occupancy and a height representation, and associates instances through predicted flow. I2-World~\cite{Liao_2025_I2World} uses intra-scene and inter-scene tokenization for efficient dynamic 4D forecasting. OccWorld~\cite{Zheng_2024_OccWorld} jointly forecasts future occupancy and ego trajectories from discrete scene tokens with a spatiotemporal generative Transformer. DOME~\cite{Gu_2024_DOME} uses trajectory-conditioned spatiotemporal diffusion to generate future scenes. PreWorld~\cite{Li_2025_PreWorld} combines 2D rendering pretraining with 3D occupancy supervision to recurrently predict future occupancy and ego trajectories.


Recent studies apply sparse scene representations to future forecasting~\cite{Dang_2026_SparseWorld,Zhu_2026_DLWM,Feng_2025_GWM,Diehl_2025_DIO}. SparseWorld~\cite{Dang_2026_SparseWorld} represents scenes with dynamic sparse queries and predicts future occupancy through ego-state-conditioned updates and position regression. DLWM~\cite{Zhu_2026_DLWM} represents scenes with sparse 3D Gaussians and learns Gaussian-flow-guided latent dynamics for future occupancy forecasting. However, recurrent sparse models may gradually lose local geometric and semantic information from earlier observations as scene states evolve. MemAnchorWorld addresses this limitation by retrieving memory anchors from historical sparse queries conditioned on the target forecast time and spatial location, then refining future occupancy predictions through residual corrections jointly gated by task and forecast time.


\section{Methodology}
\label{sec:method}

\subsection{Preliminaries}
\label{sec:preliminaries}

\subsubsection{4D Semantic Occupancy Forecasting}
Let $\mathcal{I}_{-L+1:0}=\{I_j^{(n)}\mid -L+1\le j\le0,\ 1\le n\le N_{\mathrm{cam}}\}$ denote synchronized images from $N_{\mathrm{cam}}$ cameras over $L$ observed frames, and let $\mathcal{E}_{-L+1:0}$ contain the corresponding ego poses and motion states.

Given these observations, the model predicts semantic occupancy grids in the current ego frame at future horizons $h\in\mathcal{H}$:
\begin{equation}
\begin{aligned}
\widehat{\mathcal{Y}}_{\mathcal H}
&=f_{\theta}\!\left(\mathcal{I}_{-L+1:0},
\mathcal{E}_{-L+1:0}\right),\\
\widehat{\mathcal{Y}}_{\mathcal H}
&=\{\widehat Y_h\}_{h\in\mathcal H},\\
\widehat Y_h
&\in\{1,\ldots,C,\varnothing\}^{N_x\times N_y\times N_z}.
\end{aligned}
\label{eq:forecasting_task}
\end{equation}

\subsubsection{Sparse World Models}
Sparse world models represent a scene using sparse queries, each associating a shared feature with a local set of 3D points:
\begin{equation}
\begin{aligned}
Z_k&=\{(\mathbf q_{i,k},\mathbf P_{i,k})\}_{i=1}^{N_k},\\
\mathbf P_{i,k}&=\{\mathbf p_{i,r,k}\}_{r=1}^{R},\\
\mathbf q_{i,k}&\in\mathbb R^D.
\end{aligned}
\label{eq:sparse_state}
\end{equation}

During recurrent forecasting, the model progressively updates query features and associated point positions conditioned on the ego state. The point-level prediction at a future horizon of $h$ seconds is denoted by $B_h$:
\begin{equation}
\begin{aligned}
Z_{k+1}&=F_k(Z_k),\\
B_h&=G(Z_{2h})
=\{(\mathbf{p}^{0,h}_{ir},\mathbf{s}^{0,h}_{ir})\}_{i,r},
\end{aligned}
\label{eq:base_forecast}
\end{equation}
where $Z_0$ is the current scene state and $F_k$ denotes the ego-state-conditioned scene update. The superscript $0$ indicates the point positions and semantic logits output by the base decoder. Below, we describe retrieval and correction at a fixed forecast horizon $h$ and abbreviate $\mathbf q_{i,2h}$ as $\mathbf q_i$.

\subsubsection{Error Propagation in Recurrent Forecasting}
\label{sec:recurrent_error}
In recurrent sparse forecasting, historical evidence is carried by the encoded state and fixed rollout conditions. Under fixed query correspondences, we vectorize the scene state $Z_k$ and concatenate it with the ego state $\mathbf x_k^{\mathrm{ego}}$ to obtain the augmented state $\mathbf x_k=[\operatorname{vec}(Z_k),\mathbf x_k^{\mathrm{ego}}]$. Let $\widehat{\mathbf x}_{k+1}=\mathcal F_k(\widehat{\mathbf x}_k)$, $\mathbf e_k=\widehat{\mathbf x}_k-\mathbf x_k^\star$, and $\boldsymbol\epsilon_k=\mathcal F_k(\mathbf x_k^\star)-\mathbf x_{k+1}^\star$. For continuously differentiable updates and $K=2h$, the error admits the exact decomposition
\begin{equation}
\mathbf e_K=\boldsymbol{\Phi}_{K,0}\mathbf e_0
+\sum_{j=0}^{K-1}\boldsymbol{\Phi}_{K,j+1}\boldsymbol\epsilon_j,
\label{eq:recurrent_error_propagation}
\end{equation}
where $\boldsymbol{\Phi}_{K,j}$ is the product of integrated Jacobians of the augmented transition $\mathcal F_k$ from step $j$ to $K$, with $\boldsymbol{\Phi}_{K,K}=\mathbf I$. Thus, initial and intermediate errors contribute to later predictions through the subsequent transitions. Detailed formulas and proofs are provided in \begin{NoHyper}\Cref{app:error_propagation}\end{NoHyper}.

\subsection{Framework Overview}
\label{sec:overview}

\begin{figure*}[t]
\centering
\includegraphics[width=\textwidth]{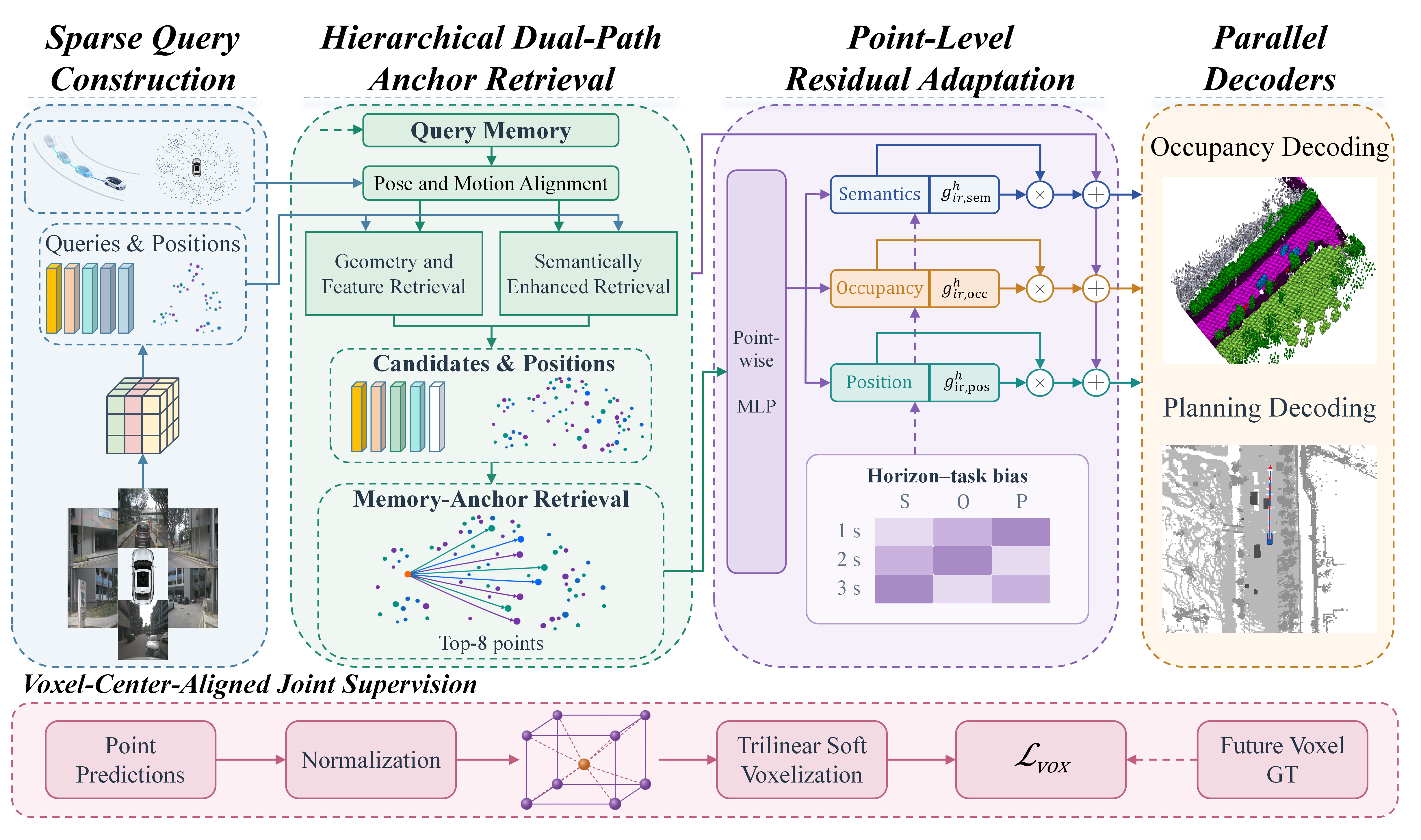}
\caption{Overview of MemAnchorWorld. First, multi-frame images generate queries and positions. Next, dual-path retrieval selects candidate queries; point-level memory-anchor retrieval selects associated points and aggregates evidence. A point-level residual adapter corrects semantic logits, occupancy confidence, and positions. Finally, decoding heads generate semantic occupancy and the ego trajectory.}
\label{fig:recallocc_framework}
\end{figure*}

As shown in \cref{fig:recallocc_framework}, MemAnchorWorld retrieves memory anchors to refine sparse future predictions and produce 4D semantic occupancy. The perception backbone first encodes surround-view images into a sparse scene state $Z_0$. Conditioned on the ego state, the forecasting backbone recurrently updates scene queries to generate predictions $B_h$ at each forecast horizon. The forecast-time-conditioned hierarchical dual-path memory-anchor retrieval module (\cref{sec:retrieval}) constructs a historical memory bank $\mathcal M$ from queries in earlier frames. Before retrieval at each forecast horizon, it uses ego poses to align historical points in a common coordinate frame, then applies object motion compensation to extrapolate them to the target forecast time. Using the corresponding scene queries and predicted points as retrieval cues and conditioned on the forecast-time embedding, the module considers historical age, semantic reliability, and geometric--semantic relevance. It first selects candidate historical queries through geometric--semantic dual-path coarse retrieval, then performs memory-anchor retrieval among their associated points and aggregates the retrieved features and semantics as correction evidence. Finally, the point-level residual adapter jointly gated by task and forecast time (\cref{sec:fusion}) fuses the retrieved evidence with the current prediction representation and forecast-time information. It generates separate residual corrections for semantics, occupancy confidence, and spatial position, with independent gates modulating the strength of each correction. Sparse occupancy decoding converts the corrected point-level outputs into semantic occupancy grids $\widehat Y_h$ at each forecast horizon.

\subsection{Hierarchical Dual-Path Anchor Retrieval}
\label{sec:retrieval}

\textbf{Memory acquisition.} 
To retain explicit access to older observations, we store their queries separately from the evolving scene state. At current time $t$, we select $K_{\mathrm{mem}}$ historical observation slots with reference times
\begin{equation}
\mathcal{T}_{t}^{\mathrm{mem}}=
\{\tau_j\}_{j=1}^{K_{\mathrm{mem}}},\qquad \tau_j<t.
\label{eq:memory_times}
\end{equation}

The selected times need not be uniformly spaced. We extract observation-state queries from these slots. Each memory entry stores a query feature, its associated points, an averaged semantic distribution, a semantic reliability score, the observation time, and the source ego pose:
\begin{equation}
m=(\mathbf{q}_m,\mathbf{P}_m,\boldsymbol{\pi}_m,
\rho_m,\tau_m,\mathbf{T}_m),
\label{eq:memory}
\end{equation}
where the mean semantic distribution $\boldsymbol{\pi}_m$ is obtained by averaging the softmax probabilities of the points associated with each query. Let $\pi_m^{(1)}\ge\pi_m^{(2)}$ be its two largest class probabilities. We compute semantic reliability $\rho_m$ as
\begin{equation}
\rho_m=\tfrac13\!\left[\pi_m^{(1)}+
(\pi_m^{(1)}-\pi_m^{(2)})+1-\frac{H(\boldsymbol{\pi}_m)}{\log C}\right].
\label{eq:reliability}
\end{equation} 

The score combines peak confidence, separation between the leading classes, and low entropy, favoring less ambiguous historical predictions.

\textbf{Historical memory alignment.}
We map historical points from their source ego frame to the current ego frame via $[\widetilde{\mathbf{p}}_{mj}^{\top},1]^{\top}=\mathbf{T}_t^{-1}\mathbf{T}_m[\mathbf{p}_{mj}^{\top},1]^{\top}$. Before candidate selection, we compensate for object motion by extrapolating the aligned points to the target forecast horizon:
\begin{equation}
\begin{aligned}
a_m^h&=t-\tau_m+h,\\
\widehat{\mathbf v}_m^{h,\mathrm{src}}
&=f_{\mathrm{motion}}\!\left(\mathbf q_m,\operatorname{PE}(a_m^h)\right),\\
\mathbf p_{mj}^{M,h}
&=\widetilde{\mathbf p}_{mj}
+a_m^h\mathbf R_t^\top\mathbf R_m\widehat{\mathbf v}_m^{h,\mathrm{src}}.
\end{aligned}
\label{eq:age}
\end{equation}

Here $a_m^h$ is the effective age, and $f_{\mathrm{motion}}$ predicts translation velocity from the historical query feature and effective-age encoding. The matrices $\mathbf R_t$ and $\mathbf R_m$ are the rotation blocks of the corresponding ego poses. Coordinate transformation and scaling by the effective age yield the displacement correction. Points associated with the same query share this translation to preserve their local configuration.

\textbf{Dual-path query selection.}
Before retrieval, we select historical queries by semantic reliability under class and spatial diversity constraints. The horizon embedding $\mathbf{t}_h$ conditions feature matching at both retrieval levels, and the sharpened historical semantic distribution is denoted $\overline{\boldsymbol{\pi}}_m$. For predicted point $(i,r)$, let $\boldsymbol{\pi}_{ir}=\operatorname{softmax}(\mathbf{s}^{0}_{ir})$. \Cref{eq:semantic_weight} reduces reliance on uncertain semantic predictions while retaining a minimum semantic contribution:
\begin{equation}
w_{ir}=\max\!\left\{\left(1-\frac{H(\boldsymbol{\pi}_{ir})}{\log C}\right)^2,\,w_{\min}\right\}.
\label{eq:semantic_weight}
\end{equation}

For query-level matching, let $\boldsymbol{\pi}_i=R^{-1}\sum_r\boldsymbol{\pi}_{ir}$ and $\overline w_i=R^{-1}\sum_r w_{ir}$. We denote the current point centroid by $\boldsymbol\mu_i$ and the motion-compensated memory centroid by $\boldsymbol\mu_m^h=R^{-1}\sum_j\mathbf p_{mj}^{M,h}$. Both paths restrict candidates to $\|\boldsymbol\mu_i-\boldsymbol\mu_m^h\|\le r_q$ before top-$k$ selection.
The coarse geometric and semantic retrieval scores are
\begin{equation}
\begin{aligned}
A_{im}^{g}&=\frac{\phi_q(\mathbf{q}_i,\mathbf{t}_h)^\top
\phi_k(\mathbf{q}_m)}{\sqrt{D}}
-\frac{\|\boldsymbol{\mu}_i-\boldsymbol{\mu}_m^h\|^2}{r_q^2}\\
&\quad-\frac{a_m^h}{a_{\max}}+\log(\rho_m+\varepsilon),\\
A_{im}^{s}&=A_{im}^{g}+\overline{w}_i
\log(\boldsymbol{\pi}_i^\top
\overline{\boldsymbol{\pi}}_m+\varepsilon),
\end{aligned}
\label{eq:coarse_scores}
\end{equation}
where $\phi_q$ and $\phi_k$ are query and key projections. Both paths combine feature similarity, spatial distance, effective age, and semantic reliability; the semantic path additionally considers semantic agreement. The age penalty favors more recent historical observations. The results from both paths are merged and deduplicated, yielding candidate historical queries and their associated point positions for subsequent point-level memory-anchor retrieval.

\textbf{Point-level memory-anchor retrieval.}
We expand candidate queries into their associated 3D points and first compute softmax-weighted contexts for geometric and additional semantic candidates. Point-level queries combine the current query feature, point position encoding, and forecast-time embedding; semantic attention heads also encode the current and historical semantic distributions. \Cref{eq:point_scores} combines point feature similarity, compensated spatial distance, effective age, semantic reliability, and semantic agreement to compute fine retrieval scores:
\begin{equation}
\begin{aligned}
b_{ir,mj}^{\ell}
&=\frac{(\mathbf{u}_{ir}^{\ell})^\top\mathbf{k}_{mj}^{\ell}}
{\sqrt{D_{\ell}}}
-\frac{\|\mathbf{p}^{0}_{ir}-\mathbf{p}_{mj}^{M,h}\|^2}{r_p^2}\\
&\quad-\frac{a_m^h}{a_{\max}}+\log(\rho_m+\varepsilon)\\
&\quad+\eta_{\ell}w_{ir}\log(
\boldsymbol{\pi}_{ir}^{\top}\overline{\boldsymbol{\pi}}_m+\varepsilon),
\end{aligned}
\label{eq:point_scores}
\end{equation}
where $\mathbf{u}_{ir}^{\ell}$ and $\mathbf{k}_{mj}^{\ell}$ are the projected point query and memory key for head $\ell$. The head width is $D_{\ell}$, and $\eta_{\ell}$ weights semantic agreement. Point-level top-$k$ selection retains only point pairs satisfying $\|\mathbf p^0_{ir}-\mathbf p^{M,h}_{mj}\|\le r_p$. Attention then aggregates evidence from the selected memory anchors according to the retrieval scores, yielding a historical feature $\mathbf{c}_{ir}$, a retrieved semantic distribution $\boldsymbol{\pi}^{M}_{ir}$, and support statistics. Position-conditioned attention lets different prediction points within the same query retrieve their own local historical evidence from memory.

\subsection{Point-Level Residual Adaptation}
\label{sec:fusion}

The usefulness of historical evidence can differ across tasks and forecast times. \Cref{eq:fusion_feature} fuses the current query representation, evidence from memory anchors, semantic agreement, and forecast-time information to form a point-level adaptation feature:
\begin{equation}
\begin{aligned}
\mathbf{f}_{ir}^{h}=\mathrm{MLP}\big([&\mathbf{q}_i,\mathrm{PE}(\mathbf{p}^{0}_{ir}),
\mathbf{c}_{ir},\mathbf{q}_i-\mathbf{c}_{ir},\mathbf{t}_h,\\
&\boldsymbol{\pi}_{ir},\boldsymbol{\pi}^{M}_{ir},
\boldsymbol{\pi}_{ir}-\boldsymbol{\pi}^{M}_{ir},\mathbf{d}_{ir}]\big).
\end{aligned}
\label{eq:fusion_feature}
\end{equation}

The statistics $\mathbf d_{ir}$ summarize support counts, reliability, motion-compensated distance, semantic uncertainty and agreement, and effective age. For points with valid historical support, let $\alpha_{ir,mj}^h$ denote attention weights normalized within each head and averaged over valid heads. We compute $\bar a_{ir}^h=\sum_{m,j}\alpha_{ir,mj}^h a_m^h$ and include $\operatorname{clip}(\bar a_{ir}^h/(a_{\max}+\varepsilon),0,1)$ in $\mathbf d_{ir}$. This age statistic retains the forecast offset $h$ and, together with $\mathbf t_h$, conditions both residuals and gates.

The adapter predicts separate residuals for semantics, occupancy confidence, and spatial position. \Cref{eq:gates} assigns an independent gate to each task with a learnable $3\times3$ forecast-time--task bias:
\begin{equation}
g_{ir,u}^{h}=v_{ir}\,\sigma\!\left(
\mathbf{w}_u^\top\mathbf{f}_{ir}^{h}+b_u+\beta_{h,u}\right),
\label{eq:gates}
\end{equation}
where $v_{ir}$ indicates whether valid point-level historical support is available. \Cref{eq:correction} adds the gated residuals to the original semantic logits, occupancy confidence, and point coordinates, respectively:
\begin{equation}
\begin{aligned}
\widehat{\mathbf{s}}_{ir}&=\mathbf{s}^{0}_{ir}
+g_{ir,\mathrm{sem}}^{h}\Delta\mathbf{s}_{ir},\\
\widehat{o}_{ir}&=\max_c s^{0}_{ir,c}
+g_{ir,\mathrm{occ}}^{h}\Delta o_{ir},\\
\widehat{\mathbf{p}}_{ir}&=\mathbf{p}^{0}_{ir}
+g_{ir,\mathrm{pos}}^{h}\Delta\mathbf{p}_{ir}.
\end{aligned}
\label{eq:correction}
\end{equation}


For supervision, we flatten $(i,r)$ into point index $i$. Corrected points inside nonempty ground-truth voxels or near their centers form positives $\mathcal P_h$ with semantic labels $c_i$ from the corresponding ground-truth voxels; other valid points form negatives $\mathcal N_h$, with $\mathcal V_h=\mathcal P_h\cup\mathcal N_h$.

With class weights $\alpha_c$, cross-entropy supervises the semantic residual and its gate on positive points,
\begin{equation}
\mathcal{L}_{\mathrm{sem}}^h=
\frac{\sum_{i\in\mathcal P_h}\alpha_{c_i}
\operatorname{CE}(\widehat{\mathbf s}_i,c_i)}
{\sum_{i\in\mathcal P_h}\alpha_{c_i}}.
\label{eq:semantic_ce}
\end{equation}

Balanced binary cross-entropy supervises the occupancy residual and its gate,
\begin{equation}
\begin{aligned}
\ell_i^{\mathrm{occ}}&=\operatorname{BCEWithLogits}(\widehat o_i,z_i),\\
\mathcal L_{\mathrm{occ}}^h&=\tfrac12\Big[
(1-\gamma_h)\underset{i\in\mathcal P_h}{\operatorname{Avg}}\ell_i^{\mathrm{occ}}\\
&\quad+\gamma_h\underset{c\in\mathcal C_h^+}{\operatorname{Avg}}
\underset{i\in\mathcal P_{h,c}}{\operatorname{Avg}}\ell_i^{\mathrm{occ}}
+\underset{i\in\mathcal N_h}{\operatorname{Avg}}\ell_i^{\mathrm{occ}}\Big].
\end{aligned}
\label{eq:balanced_occ}
\end{equation}
Here $z_i$ is $1$ for positives and $0$ otherwise, $\mathcal P_{h,c}$ contains positives of class $c\in\mathcal C_h^+$, and $\gamma_h$ controls class balancing.

An anchor regularizer limits occupancy-logit deviations from the stop-gradient ($\operatorname{sg}$) backbone confidence proxy,
\begin{equation}
\mathcal L_{\mathrm{anc}}^h=
\underset{i\in\mathcal V_h}{\operatorname{Avg}}\,
\operatorname{SmoothL1}\!\left(
\widehat o_i,\operatorname{sg}\!\left(\max_c s^0_{i,c}\right)\right).
\label{eq:anchor_loss}
\end{equation}

A threshold-margin loss aligns occupancy correction and gating with decoding. Let $\kappa_c=\log(\theta_c/(1-\theta_c))$ for class probability threshold $\theta_c$. With margin $\delta$,
\begin{equation}
\ell_i^{\mathrm{thr}}=
\begin{cases}
\operatorname{softplus}(\kappa_{c_i}+\delta-\widehat o_i),&i\in\mathcal P_h,\\
\operatorname{softplus}(\widehat o_i-\kappa_{\widehat c_i}+\delta),&i\in\mathcal N_h,
\end{cases}
\label{eq:threshold_loss}
\end{equation}
where $\widehat c_i=\arg\max_c\widehat s_{i,c}$. The loss $\mathcal L_{\mathrm{thr}}^h$ averages both group means, pushing positives above their true-class thresholds and negatives below their predicted-class thresholds. The total objective and remaining losses are detailed in \begin{NoHyper}\Cref{app:loss_details}\end{NoHyper}.


\section{Experiments}
\label{sec:experiments}

\begin{table*}[!t]
\centering
\caption{4D occupancy forecasting on Occ3D-nuScenes. All models use camera input. Baseline values, including averages, are taken from the corresponding publications. Metric values are percentages; parentheses indicate gains over SparseWorld in percentage points.}
\label{tab:occupancy_forecasting}
\setlength{\tabcolsep}{0.5pt}
\small
\begin{tabularx}{\textwidth}{lc*{8}{>{\centering\arraybackslash}X}}
\toprule
\multirow{2}{*}{\textbf{Method}} & \multirow{2}{*}{\textbf{Aux. Sup.}}
& \multicolumn{4}{c}{\textbf{mIoU} $\uparrow$} & \multicolumn{4}{c}{\textbf{IoU} $\uparrow$} \\
\cmidrule(lr){3-6}\cmidrule(lr){7-10}
& & \textbf{1s} & \textbf{2s} & \textbf{3s} & \textbf{Avg.} & \textbf{1s} & \textbf{2s} & \textbf{3s} & \textbf{Avg.} \\
\midrule
OccWorld-S~\cite{Zheng_2024_OccWorld} & None & 0.28 & 0.26 & 0.24 & 0.26 & 5.05 & 5.01 & 4.95 & 5.00 \\
RenderWorld~\cite{Yan_2024_RenderWorld} & None & 2.83 & 2.55 & 2.37 & 2.58 & 14.61 & 13.61 & 12.98 & 13.73 \\
OccWorld-T~\cite{Zheng_2024_OccWorld} & \begin{tabular}[c]{@{}c@{}}Semantic\\LiDAR\end{tabular} & 4.68 & 3.36 & 2.63 & 3.56 & 9.32 & 8.23 & 7.47 & 8.34 \\
\midrule
OccWorld-D~\cite{Zheng_2024_OccWorld} & 3D Occ & 11.55 & 8.10 & 6.22 & 8.62 & 18.90 & 16.26 & 14.43 & 16.53 \\
PreWorld~\cite{Li_2025_PreWorld} & 3D Occ & 11.69 & 8.72 & 6.77 & 9.06 & 23.01 & 20.79 & 18.84 & 20.88 \\
PreWorld~\cite{Li_2025_PreWorld} & 2D \& 3D Occ & 12.27 & 9.24 & 7.15 & 9.55 & \underline{23.62} & 21.62 & 19.63 & 21.62 \\
DFIT-OccWorld~\cite{Zhang_2024_DFITOccWorld} & 3D Occ & 13.38 & 10.16 & 7.96 & 10.50 & 19.18 & 16.85 & 15.02 & 17.02 \\
SparseWorld~\cite{Dang_2026_SparseWorld} & 3D Occ & \underline{14.93} & \underline{13.15} & \underline{11.51} & \underline{13.20} & 22.96 & \underline{22.10} & \underline{21.05} & \underline{22.03} \\
\specialrule{\lightrulewidth}{\aboverulesep}{0pt}
\rowcolor{MemAnchorWorldRow}
\multicolumn{1}{c}{\textbf{MemAnchorWorld}} & 3D Occ & \textbf{15.39}\,{\scriptsize(\textcolor{red}{$\uparrow$0.46})} & \textbf{13.40}\,{\scriptsize(\textcolor{red}{$\uparrow$0.25})} & \textbf{11.87}\,{\scriptsize(\textcolor{red}{$\uparrow$0.36})} & \textbf{13.55}\,{\scriptsize(\textcolor{red}{$\uparrow$0.35})} & \textbf{23.88}\,{\scriptsize(\textcolor{red}{$\uparrow$0.92})} & \textbf{22.86}\,{\scriptsize(\textcolor{red}{$\uparrow$0.76})} & \textbf{21.78}\,{\scriptsize(\textcolor{red}{$\uparrow$0.73})} & \textbf{22.84}\,{\scriptsize(\textcolor{red}{$\uparrow$0.81})} \\
\specialrule{\heavyrulewidth}{0pt}{0pt}
\end{tabularx}
\end{table*}

\begin{table}[!t]
\centering
\caption{Motion planning on Occ3D-nuScenes. The two metric groups report $L_2$ error and box collision rate, respectively. Baseline values follow the corresponding publications.}
\label{tab:motion_planning}
\setlength{\tabcolsep}{2pt}
\small
\resizebox{\columnwidth}{!}{%
\begin{tabular}{*{10}{c}}
\toprule
\multirow{2}{*}{\textbf{Method}} & \multirow{2}{*}{\textbf{Aux. Sup.}}
& \multicolumn{4}{c}{\textbf{$L_2$ (m)} $\downarrow$} & \multicolumn{4}{c}{\textbf{Box Collision (\%)} $\downarrow$} \\
\cmidrule(lr){3-6}\cmidrule(lr){7-10}
& & \textbf{1s} & \textbf{2s} & \textbf{3s} & \textbf{Avg.} & \textbf{1s} & \textbf{2s} & \textbf{3s} & \textbf{Avg.} \\
\midrule
UniAD~\cite{Hu_2023_CVPR} & \begin{tabular}[c]{@{}c@{}}Map, Box,\\Motion, Track,\\3D Occ\end{tabular} & 0.48 & 0.96 & 1.65 & 1.03 & \textbf{0.05} & \underline{0.17} & 0.71 & 0.31 \\
VAD~\cite{Jiang_2023_ICCV} & \begin{tabular}[c]{@{}c@{}}Map, Box,\\Motion\end{tabular} & 0.54 & 1.15 & 1.98 & 1.22 & 0.10 & 0.24 & 0.96 & 0.43 \\
OccNet~\cite{Tong_2023_OccNet} & \begin{tabular}[c]{@{}c@{}}Map, Box,\\3D Occ\end{tabular} & 1.29 & 2.13 & 2.99 & 2.14 & 0.21 & 0.59 & 1.37 & 0.72 \\
OccWorld-D~\cite{Zheng_2024_OccWorld} & 3D Occ & 0.52 & 1.27 & 2.41 & 1.40 & 0.12 & 0.40 & 2.08 & 0.87 \\
RenderWorld-V~\cite{Yan_2024_RenderWorld} & 3D Occ & 0.48 & 1.30 & 2.67 & 1.48 & 0.14 & 0.55 & 2.23 & 0.97 \\
PreWorld~\cite{Li_2025_PreWorld} & \begin{tabular}[c]{@{}c@{}}2D \&\\3D Occ\end{tabular} & 0.22 & 0.30 & 0.40 & 0.31 & 0.21 & 0.66 & 0.71 & 0.53 \\
SparseWorld~\cite{Dang_2026_SparseWorld} & 3D Occ & \underline{0.19} & \underline{0.25} & \underline{0.36} & \underline{0.27} & 0.11 & 0.29 & \underline{0.46} & \underline{0.29} \\
\specialrule{\lightrulewidth}{\aboverulesep}{0pt}
\rowcolor{MemAnchorWorldRow}
\multicolumn{1}{c}{\textbf{MemAnchorWorld}} & 3D Occ & \textbf{0.16} & \textbf{0.19} & \textbf{0.24} & \textbf{0.20} & \underline{0.09} & \textbf{0.10} & \textbf{0.11} & \textbf{0.10} \\
\specialrule{\heavyrulewidth}{0pt}{0pt}
\end{tabular}
}
\end{table}

\subsection{Experimental Details}
\label{sec:experimental_details}

\textbf{Dataset.} We conduct experiments on Occ3D-nuScenes~\cite{Tian_2023_Occ3D}, which provides visibility-aware semantic occupancy annotations for nuScenes~\cite{Caesar_2020_nuScenes}. Following the forecasting protocol of sparse world models, we use 700 scenes for training and 150 for validation. Each scene lasts approximately 20\,s, with annotated frames at 0.5\,s intervals. The evaluation region spans $[-40,40]$\,m along both horizontal axes and $[-1,5.4]$\,m vertically. It is discretized into a $200\times200\times16$ grid with 0.4\,m voxels and 17 occupied semantic classes plus free space. The forecasting backbone observes the current frame and four preceding frames over 2\,s, then rolls out six future states at 0.5\,s intervals. We evaluate the predictions at 1, 2, and 3\,s.

\textbf{Implementation Details.}
We use a pretrained SparseWorld-R50 backbone~\cite{Dang_2026_SparseWorld}, comprising a ResNet-50 image encoder~\cite{He_2016_ResNet} and an FPN neck~\cite{Lin_2017_FPN} for six surround-view images at $256\times704$ resolution. Its six decoding layers expand each query into $(1,4,16,24,32,48)$ points, and its query schedule assigns 720 queries to the current state and $(60,60,60,60,40,40)$ to the six future steps. We freeze the perception and forecasting backbone while optimizing MemAnchorWorld's memory retrieval and residual-adaptation modules with the objective in \begin{NoHyper}\Cref{app:loss_details}\end{NoHyper}, \begin{NoHyper}\cref{eq:total_loss}\end{NoHyper}.

In this implementation, $K_{\mathrm{mem}}=3$ historical slots are centered at $t-2.5$, $t-3.5$, and $t-4.5$\,s, with a 0.35\,s matching tolerance and a fixed effective-age limit $a_{\max}=8$\,s. Each slot contributes at most 256 queries, giving a memory of at most 768 entries. Both retrieval levels use a 6\,m radius and retain up to eight candidates per path or point-level head; implementation details are provided in \begin{NoHyper}\Cref{app:retrieval_details}\end{NoHyper}.

\textbf{Evaluation Metrics.}
For occupancy forecasting, we report semantic mean intersection-over-union (mIoU) over the 17 occupied classes and class-agnostic occupied-space IoU at 1, 2, and 3 seconds. For motion planning, we report the Euclidean $L_2$ displacement error  and the box collision rate at the same horizons, together with their averages.

\begin{figure*}[!t]
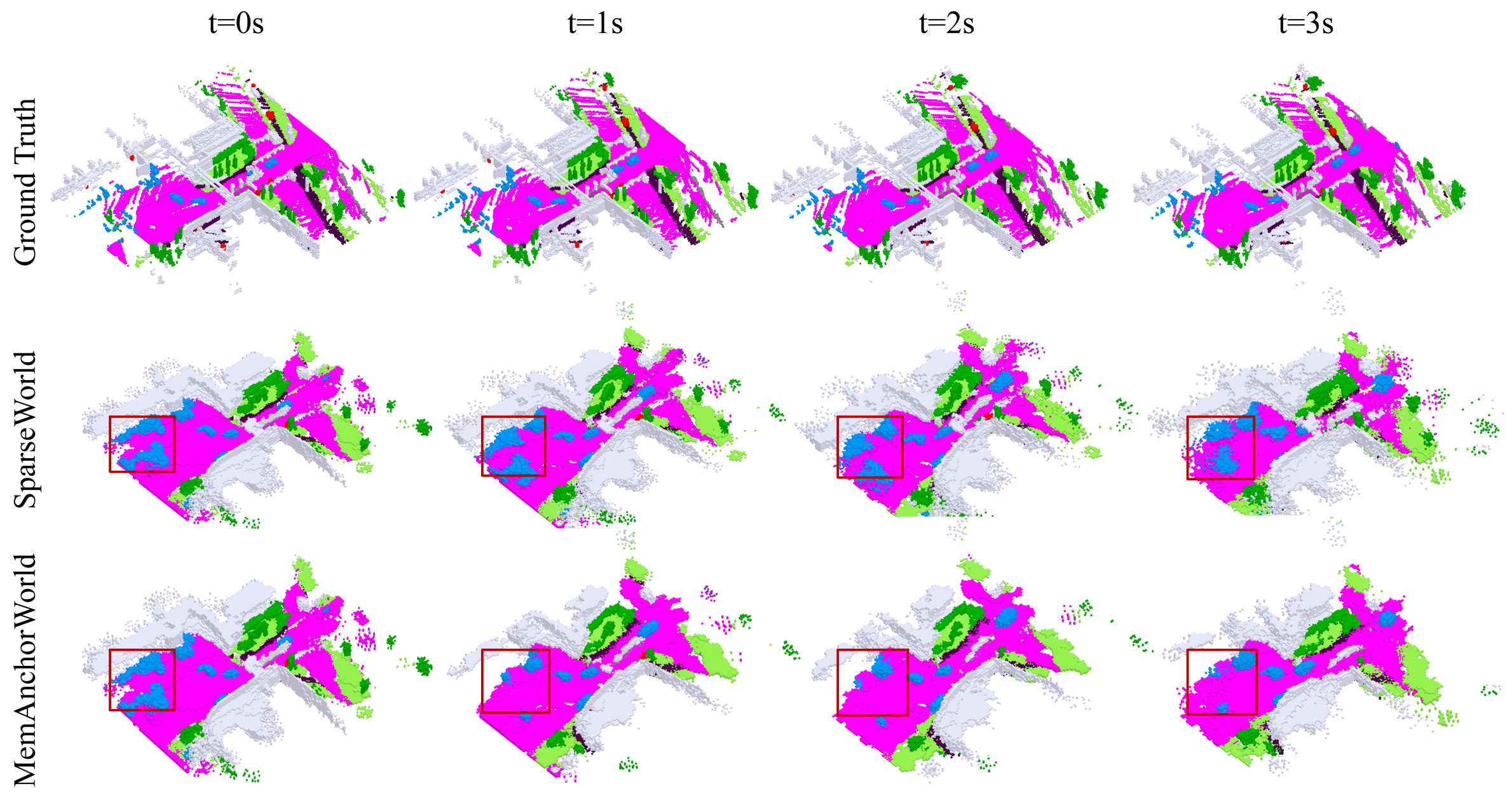

\centering
\MemAnchorQualitative
\caption{Qualitative 4D semantic occupancy forecasting on Occ3D-nuScenes. Columns show the current frame and forecasts at 1, 2, and 3\,s. From top to bottom, rows show ground truth, SparseWorld, and MemAnchorWorld. Red boxes highlight local regions where MemAnchorWorld better predicts semantic occupancy and its spatial support over time.}
\label{fig:recallocc_qualitative}
\end{figure*}

\subsection{Main Results}
\label{sec:main_results}

\phantomsection
\textbf{4D Occupancy Forecasting.}
\label{sec:forecasting_results}
\Cref{tab:occupancy_forecasting} compares MemAnchorWorld with prior occupancy world models. For trajectory conditioning, OccWorld, RenderWorld, PreWorld, SparseWorld, and MemAnchorWorld use observed ego motion~\cite{Zheng_2024_OccWorld,Yan_2024_RenderWorld,Li_2025_PreWorld,Dang_2026_SparseWorld}; future ground-truth ego trajectories serve as training or evaluation targets rather than inference inputs. MemAnchorWorld improves both mIoU and IoU over SparseWorld at every evaluated horizon. Its average mIoU and IoU reach 13.55\% and 22.84\%, exceeding SparseWorld by 0.35 and 0.81 percentage points, respectively. MemAnchorWorld achieves the highest mIoU and IoU at all evaluated horizons and on average among the compared methods. These results highlight MemAnchorWorld's performance advantages in future occupancy forecasting and its ability to predict future semantic and spatial occupancy states more accurately.

\phantomsection
\textbf{Motion Planning.}
\label{sec:planning_results}
\Cref{tab:motion_planning} compares $L_2$ displacement error and box collision rate at 1, 2, and 3\,s, together with their averages. MemAnchorWorld achieves the lowest average $L_2$ error and average box collision rate among the compared methods. Additionally, Obj Collision, the trajectory-reference-point collision rate after GT-collision masking, is 0\% at all three horizons. Metric definitions and the evaluation are detailed in \begin{NoHyper}\Cref{app:planning_evaluation}\end{NoHyper}. These results suggest that MemAnchorWorld effectively captures the end-to-end interactions among occupancy representation, trajectory prediction, and planning across all three forecast horizons.

\subsection{Mechanism Diagnosis}
\label{sec:mechanism_diagnosis}

\begin{table}[!ht]
\centering
\caption{Historical-memory diagnosis on points with valid retrieved support whose uncorrected positions lie in occupied ground-truth voxels.}
\label{tab:mechanism_diagnosis}
\setlength{\tabcolsep}{3pt}
\small
\begin{tabular}{@{}cccc@{}}
\toprule
\textbf{Horizon} & \textbf{Hist. compl.} & \textbf{WC correction} & \textbf{CW retention} \\
\midrule
1s & 25.25\% & 9.57\% & 97.47\% \\
2s & 24.29\% & 11.12\% & 97.19\% \\
3s & 23.19\% & 12.58\% & 96.64\% \\
\bottomrule
\end{tabular}
\end{table}

At 1, 2, and 3\,s, we evaluate historical complementarity (Hist. compl.), correction of wrong base predictions with correct history (WC), and retention of correct base predictions with wrong history (CW). Metric definitions are provided in \begin{NoHyper}\Cref{app:mechanism_counts}\end{NoHyper}.

\textbf{Complementary Historical Evidence.}
\Cref{tab:mechanism_diagnosis} reports rates of 25.25\%, 24.29\%, and 23.19\% at the three horizons, respectively. Thus, roughly one quarter of points misclassified by the base prediction can retrieve semantically correct memory anchors from history, providing complementary semantic evidence for refining the corresponding future point-level predictions.

\textbf{Use of Complementary History.}
Among points with an incorrect base prediction but a correct historical class, the adapter corrects 9.57\%, 11.12\%, and 12.58\% at 1, 2, and 3\,s, respectively. The correction rate increases with the forecast horizon. These results show that the adapter can use correct retrieved historical semantics to guide residual corrections, recovering the correct class for some initially misclassified points and translating complementary evidence into effective prediction refinement.

\textbf{Resistance to Conflicting History.}
When the base prediction is correct but the historical class is wrong, 97.47\%, 97.19\%, and 96.64\% of points remain correct after adaptation at 1, 2, and 3\,s, respectively. These results support the effectiveness of the gated adapter in selectively fusing historical evidence, limiting interference from erroneous history while preserving initially correct point-level predictions under semantic conflicts.

\begin{figure*}[!t]
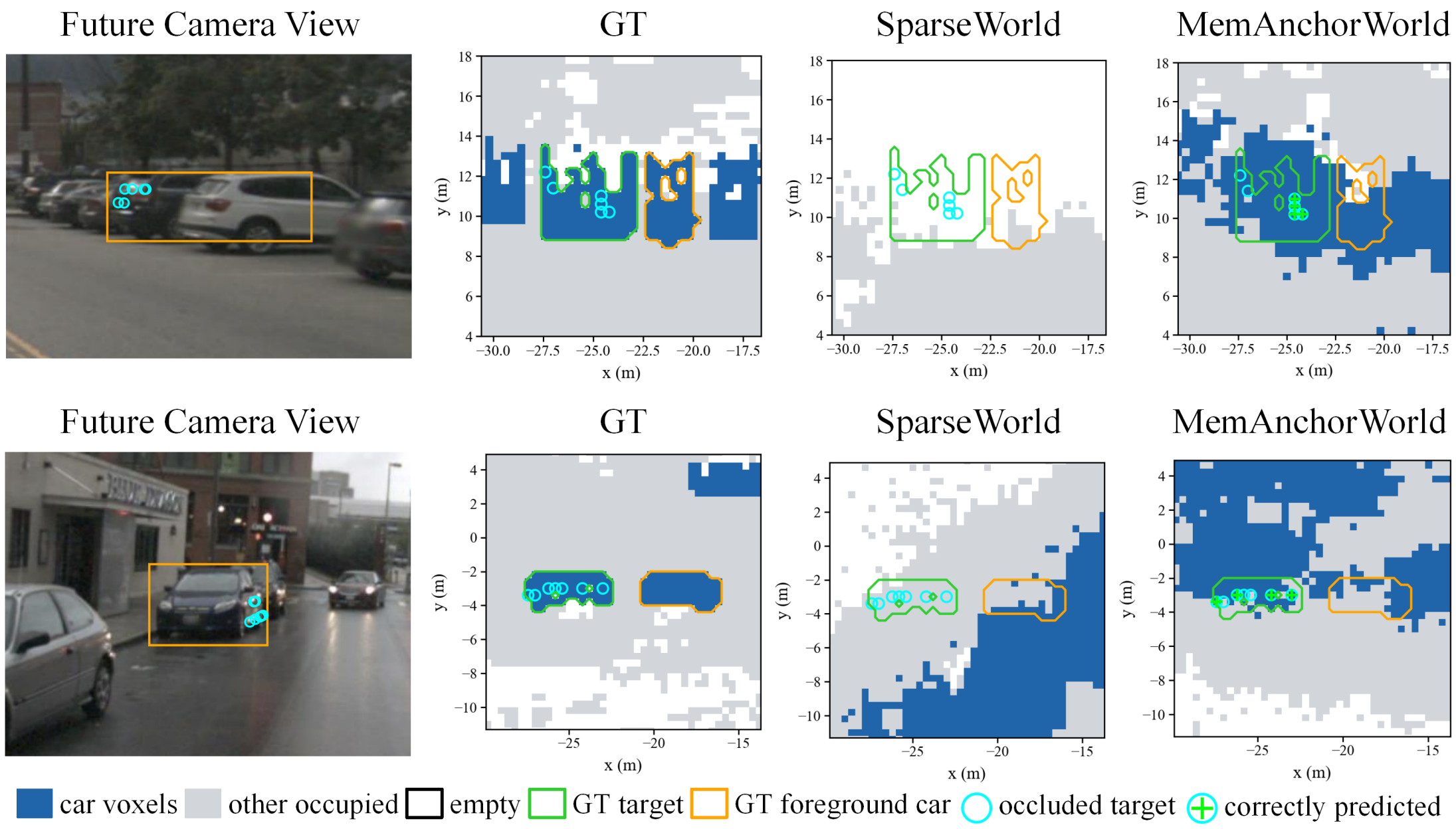

\centering
\MemAnchorOcclusion
\caption{Vehicle-occlusion forecasts at 1\,s in two scenes. Columns show the future camera view, ground truth, SparseWorld, and MemAnchorWorld. Green and orange outlines mark the target and foreground vehicles, respectively.}
\label{fig:recallocc_occlusion}
\end{figure*}

\subsection{Ablation Studies}
\label{sec:ablation_studies}


\Cref{tab:memory_ablation} reports ablation results for the core components using average mIoU and IoU over the 1--3\,s forecasts. SparseWorld serves as our baseline, achieving average mIoU and IoU scores of 13.20\% and 22.03\%, respectively. When only memory-anchor retrieval is added, ordinary feature fusion replaces gated fusion. Average mIoU and IoU improve over the baseline by 0.04 and 0.22 percentage points, respectively, indicating that retrieved memory anchors provide historical evidence that improves future predictions.

When only task--horizon gated adaptation is added, an MLP encodes queries at the target forecast horizon in place of the memory-anchor retrieval branch. Average mIoU and IoU improve over the baseline by 0.11 and 0.31 percentage points, respectively. These gains indicate that task--horizon gated adaptation can use the current prediction representation for adaptive correction, reducing the impact of error propagation and accumulation on future predictions. With both components fully enabled, average mIoU and IoU reach 13.55\% and 22.84\%, outperforming both single-component variants and indicating that memory-anchor retrieval and task--horizon gated adaptation are complementary and jointly improve future occupancy forecasting.

\subsection{Visualizations}
\label{sec:visualizations}

\textbf{4D Semantic Occupancy Forecasting.}
As shown in \cref{fig:recallocc_qualitative}, MemAnchorWorld and SparseWorld produce similar occupancy at 0\,s because they share the perception backbone. At 1--3\,s, SparseWorld misclassifies some road regions within the red boxes as vehicles, producing false vehicle occupancy, whereas MemAnchorWorld more accurately predicts the semantic classes of the regions around vehicles. These results suggest that memory anchors retrieved by MemAnchorWorld from historical memory supply class cues for misclassified regions, enabling gated corrections to reduce semantic errors in future predictions.

\begin{table}[!t]
\centering
\caption{Ablation studies of core model components.}
\label{tab:memory_ablation}
\setlength{\tabcolsep}{3pt}
\small
\begin{tabular}{@{}cccc@{}}
\toprule
\begin{tabular}[c]{@{}c@{}}\textbf{Memory}\\\textbf{retrieval}\end{tabular} &
\begin{tabular}[c]{@{}c@{}}\textbf{Task--horizon}\\\textbf{gating}\end{tabular} &
\textbf{Avg. mIoU} $\uparrow$ & \textbf{Avg. IoU} $\uparrow$ \\
\midrule
$-$ & $-$ & 13.20 & 22.03 \\
$-$ & $\checkmark$ & 13.31 & 22.34 \\
$\checkmark$ & $-$ & 13.24 & 22.25 \\
\midrule
$\checkmark$ & $\checkmark$ & 13.55 & 22.84 \\
\bottomrule
\end{tabular}
\end{table}

\textbf{Effectiveness Evaluation.}
\Cref{fig:recallocc_occlusion} shows predictions for two occlusion scenes. The two scenes contain six and seven occluded target voxels, respectively. SparseWorld correctly predicts none of these target voxels in either scene, whereas MemAnchorWorld correctly predicts four target voxels in each. Vehicle IoU within the fixed local 3D regions improves from zero to 19.94\% and 11.82\%. These examples show that MemAnchorWorld can use evidence from earlier observations to supplement incomplete scene representations, demonstrating the effectiveness of its memory-anchor retrieval and fusion mechanisms.

\section{Conclusion}
\label{sec:conclusion}

In this paper, we presented MemAnchorWorld, a sparse occupancy world model addressing the temporal-evidence bottleneck in long-horizon forecasting. The model gathers historical evidence through hierarchical dual-path memory-anchor retrieval and uses a residual adapter jointly gated by task and forecast time to correct semantics, occupancy confidence, and spatial positions. Experiments on Occ3D-nuScenes demonstrate its forecasting and planning performance. Our retrieval does not explicitly establish object-level correspondences across time; future work will explore stronger motion-aware associations to improve the reliability of memory-anchor retrieval in dynamic scenes.
\section*{LLM Usage Statement}
We used GPT6.1 (OpenAI) to assist with manuscript translation, language polishing and grammar checking, literature lookup and reference formatting, mathematical exposition, and the preparation of LaTeX source, tables, and figure layouts. All reported experimental measurements were obtained through the training and evaluation pipelines described in this work. The authors reviewed and verified all AI-assisted content and take full responsibility for the final manuscript, conclusions, and experimental results.
\par
{
    \small
    \bibliographystyle{ieeenat_fullname}
    \bibliography{main}
}


\end{document}